\documentclass[journal]{IEEEtran}
\usepackage[T1]{fontenc}% optional T1 font encoding

\usepackage{cite}
\usepackage{amsmath}
\usepackage{amsthm}
\usepackage{amssymb}
\usepackage{dsfont}
\usepackage{graphicx}
\usepackage{multicol}
\usepackage{textcomp}
\usepackage{xcolor}
\usepackage{xspace}
\usepackage{bbm}
\usepackage{algorithm,algorithmic}
\usepackage{hyperref}
\usepackage{tabularray}
\usepackage{booktabs}
\usepackage{wrapfig}
\usepackage{booktabs}
\usepackage{makecell}
\usepackage{multirow}
\usepackage{booktabs}

\usepackage{color,soul}
\definecolor{lightgray}{gray}{.9}

\usepackage{etoolbox}
\usepackage[normalem]{ulem}

\usepackage{pifont}
\newcommand{\cmark}{\ding{51}}%
\newcommand{\xmark}{\ding{55}}%
\usepackage{subfigure}
\DeclareMathAlphabet{\mathcal}{OMS}{cmsy}{m}{n}
\DeclareMathOperator*{\sign}{sign}
\makeatletter
\patchcmd{\@makecaption}
{\scshape}
{}
{}
{}
\makeatother

\usepackage[cmintegrals]{newtxmath}

\makeatletter
\newcommand\fs@betterruled{%
  \def\@fs@cfont{\bfseries}\let\@fs@capt\floatc@ruled
  \def\@fs@pre{\vspace*{5pt}\hrule height.8pt depth0pt \kern2pt}%
  \def\@fs@post{\kern2pt\hrule\relax}%
  \def\@fs@mid{\kern2pt\hrule\kern2pt}%
  \let\@fs@iftopcapt\iftrue}
\floatstyle{betterruled}
\restylefloat{algorithm}
\makeatother

\usepackage{fancyhdr}

\begin{document}

\title{Towards Communication-Efficient Adversarial Federated Learning for Robust Edge Intelligence}

\author{Yu Qiao,
        Apurba Adhikary,
        Huy Q. Le,
        Eui-Nam Huh,~\IEEEmembership{Member, IEEE}, \\
        Zhu Han,~\IEEEmembership{Fellow, IEEE},
        and Choong Seon Hong,~\IEEEmembership{Fellow, IEEE}% <-this % stops a space
\IEEEcompsocitemizethanks{
\IEEEcompsocthanksitem Yu Qiao, Huy Q. Le, Eui-Nam Huh, and Choong Seon Hong are with the School of Computing, Kyung Hee University, Yongin-si 17104, Republic of Korea (e-mail: qiaoyu@khu.ac.kr; quanghuy69@khu.ac.kr; johnhuh@khu.ac.kr; cshong@khu.ac.kr).
\IEEEcompsocthanksitem{Apurba Adhikary is with the Department of Computer Science and Engineering, School of Computing, Kyung Hee University, Yongin-si 17104, Republic of Korea, and also with the Department of Information and Communication Engineering, Noakhali Science and Technology University, Noakhali-3814, Bangladesh (e-mail: apurba@khu.ac.kr).}
\IEEEcompsocthanksitem Zhu Han is with the Department of Electrical and Computer Engineering, University of Houston, Houston, TX 77004 USA, and also with the Department of Computer Science and Engineering, Kyung Hee University, Seoul 446-701, South Korea (e-mail: hanzhu22@gmail.com).

Corresponding author: Choong Seon Hong}}

\maketitle

\begin{abstract}
Federated learning (FL) has gained significant attention for enabling decentralized training on edge networks without exposing raw data. However, FL models remain susceptible to adversarial attacks and performance degradation in non-IID data settings, thus posing challenges to both robustness and accuracy. This paper aims to achieve communication-efficient adversarial federated learning (AFL) by leveraging a pre-trained model to enhance both robustness and accuracy under adversarial attacks and non-IID challenges in AFL. By leveraging the concise knowledge embedded in the class probabilities from a pre-trained model for both clean and adversarial images, we propose a pre-trained model-guided adversarial federated learning (PM-AFL) framework. This framework integrates vanilla and adversarial mixture knowledge distillation to effectively balance accuracy and robustness while promoting local models to learn from diverse data. Specifically, for clean accuracy, we adopt a dual distillation strategy where the class probabilities of randomly paired images, and their blended versions are aligned between the teacher model and the local models. For adversarial robustness, we employ a similar distillation approach but replace clean samples on the local side with adversarial examples. Moreover, by considering the bias between local and global models, we also incorporate a consistency regularization term to ensure that local adversarial predictions stay aligned with their corresponding global clean ones. These strategies collectively enable local models to absorb diverse knowledge from the teacher model while maintaining close alignment with the global model, thereby mitigating overfitting to local optima and enhancing the generalization of the global model. Experiments demonstrate that the PM-AFL-based framework not only significantly outperforms other methods but also maintains communication efficiency.
\end{abstract}


\IEEEpeerreviewmaketitle

\section{Introduction}
\label{sec:intro}
\IEEEPARstart{N}{owadays}, advancements in deep learning, increased computational power, and the vast amounts of data available on the Internet have driven the emergence of large language models (LLMs)~\cite{achiam2023gpt,team2023gemini,qiao2025deepseek}. These models have demonstrated remarkable capabilities in a wide range of tasks, including human-like conversations~\cite{torne2024chatgpt}, image and text generation~\cite{lin2024evaluating}, and information retrieval~\cite{zhang2023empirical}. Meanwhile, billions of devices in edge networks, such as smartphones, IoT devices, and autonomous vehicles, generate vast amounts of data daily~\cite{qiao2023mp}, which provides an exceptionally rich source for enhancing LLMs~\cite{qiao2025deepseek}. However, despite their superior performance in natural language processing and other AI tasks, concerns have arisen regarding the legality of the data used to train these models. In addition, due to privacy concerns, data owners in edge networks may be reluctant to share their data, thus leading to the issue of data silos. To address these challenges, federated learning (FL)~\cite{mcmahan2017communication} has emerged as an advanced paradigm for training machine learning models in a decentralized manner. In FL, multiple clients collaborate to build a shared global model while keeping their private data confidential~\cite{mcmahan2017communication}. This approach is particularly relevant in scenarios involving sensitive or personal information such as healthcare~\cite{antunes2022federated}, finance~\cite{long2020federated}, and social media~\cite{salim2022perturbation}, as it preserves data privacy and security. However, despite its advantages, FL faces several challenges, notably the non-independent and identically distributed (non-IID) data issue, which can hinder model performance and generalization~\cite{qiao2023mp}.

Recently, similar to centralized machine learning,  researchers have also found that FL models are vulnerable to adversarial examples (AEs)~\cite{szegedy2013intriguing,goodfellow2014explaining}. These AEs are images subtly altered with carefully crafted, imperceptible perturbations designed to mislead model predictions. The adversarial attacks can pose a significant threat to the secure deployment of FL models in real-world applications, such as autonomous driving and medical image analysis~\cite{zizzo2020fat,hong2023federated}. Furthermore, the inherent non-IID data distribution across clients may exacerbate the threat, making it even more challenging to achieve both adversarial robustness and high natural accuracy~\cite{zhang2023delving,qiao2024logit}. To address these concerns, researchers have explored various strategies. One approach is robustness sharing~\cite{hong2023federated, zizzo2020fat}, where adversarial training (AT) is conducted at high-resource clients, and the resulting robustness is shared with low-resource clients. Another line of research focuses on logit adjustment~\cite{zhang2023delving, chen2022calfat, qiao2024noms_fedalc}, which involves reweighting the logit of adversarially trained models to improve their robustness. In addition, feature sharing~\cite{qiao2024logit, qiao2024fedccl} techniques are employed to boost the resilience of FL models against attacks by contrasting adversarial features with clean ones. Nonetheless, these methods share a common weakness: they require local clients to train their models from scratch, resulting in excessive computational and communication demands, particularly when dealing with large-scale models.

Previous research on adversarial knowledge distillation (AKD)~\cite{hinton2015distilling,beyer2022knowledge,goldblum2020adversarially,zi2021revisiting} has shown that a robust teacher model can simultaneously produce student models with higher clean and robust accuracy. However, whether this observation holds in the context of adversarial federated learning (AFL) remains an open question. To investigate this, we conduct a toy example, as shown in Table~\ref{tab:motivation}. We begin by exploring vanilla knowledge distillation (VKD), which relies solely on clean samples for distillation. The results indicate that while a significant portion of clean performance can be inherited, the transfer of robust behavior is limited (56.78\% \textit{vs.} 3.20\%). Next, we examine AKD, which uses AEs instead of clean samples for distillation. We observe that, compared to VKD, AKD exhibits improved robustness but compromises clean accuracy. Interestingly, this differs somewhat from prior findings~\cite{goldblum2020adversarially,huang2023boosting}, as we note that although AKD inherits both clean and robust accuracy to some extent, its clean accuracy is significantly lower than that of VKD (45.64\% \textit{vs.} 56.78\%). This inspires us to leverage the advantages of both VKD and AKD to strike a balance between clean accuracy and robust accuracy. On the other hand, to defend against adversarial attacks, a widely adopted approach is AT, which has proven to be an effective method for enhancing adversarial robustness~\cite{shafahi2019adversarial,bai2021recent}. For instance, as shown in Table~\ref{tab:motivation}, FedPGD~\cite{madry2018towards}, an AT-powered AFL algorithm, demonstrates a significant improvement in robust accuracy (17.22\% \textit{vs.} 0.00\%) compared to the vanilla federated algorithm, FedAvg~\cite{mcmahan2017communication}. These results highlight that the AT strategy is also effective in the context of FL. However, AT can introduce significant computational complexity and, due to the large model capacity required~\cite{goodfellow2014explaining,madry2018towards,zhang2019theoretically}, it is also communication-intensive in FL settings. For instance, FedPGD requires more communication rounds (200 \textit{vs.} 150) and consumes more communication resources (11.69M \textit{vs.} 0.30M) compared to distillation-based approaches, such as VKD~\cite{hinton2015distilling} and AKD~\cite{goldblum2020adversarially}. Finally, the communication-efficient method with a balanced clean and robust accuracy can be observed in both the PM-AFL and PM-AFL++ methods, with PM-AFL++ demonstrating better performance than PM-AFL. Note that even though we conduct experiments with more training rounds for investigation, all methods show only slight improvements in accuracy.

Building on the aforementioned findings and discussions, we are motivated to explore the strategy of pre-trained models in the context of AFL to enhance communication efficiency while balancing accuracy and robustness. We refer to this strategy as the pre-trained model-guided adversarial federated learning (PM-AFL) framework, which follows the standard FL training paradigm but enhances local updates by allowing each model to absorb knowledge from a well-generalized teacher model. This mitigates the limitations of relying solely on locally available data, thereby improving the generalization ability of the models. Moreover, leveraging the pre-trained model helps reduce communication overhead and accelerates the convergence of the AFL process, making it particularly suitable for resource-constrained environments. In this paper, we develop our proposal from two distinct perspectives. First, from the model training perspective, we introduce a dual-KD strategy that integrates both VKD and AKD processes within the PM-AFL framework to strike a balance between accuracy and robustness. Second, from the data augmentation perspective, we suggest leveraging locally augmented data within the PM-AFL framework to enhance data diversity, thereby improving the models' generalization. By integrating these two perspectives, we present PM-AFL++, a unified and enhanced framework, as our final proposal, which consists of three core components. First, to improve clean accuracy, we encourage the clean representations generated by the local model for both natural samples and their mixed counterparts to closely align with the corresponding clean representations from the teacher model. Second, to enhance adversarial robustness, we encourage the adversarial representations produced by the local model for natural samples and their mixed counterparts to align with the corresponding clean representations generated by the teacher model. Finally, to address the non-IID data challenge in FL, we further introduce a global alignment term that encourages local adversarial features to align with their corresponding global clean features, thereby mitigating the impact of non-IID data in the AFL environment. Overall, these strategies are expected to position PM-AFL++ as a competitive approach, significantly improving both the accuracy and robustness of the model, while effectively addressing the challenges posed by non-IID data in AFL.

\begin{table}[t]
\centering
\caption{Experiments are conducted on CIFAR-10 with a Dirichlet~\cite{balakrishnan2014continuous} parameter of 0.1. "Acc." and "Rob." represent clean and robust accuracy (\%), respectively, where "Rob." is evaluated using AutoAttack~\cite{croce2020reliable}.}
\vspace{-5px}
\label{tab:motivation}
\resizebox{0.98\linewidth}{!}{%
\begin{tabular}{l|c|c|c|c} 
\toprule
Setup & Param.$\downarrow$  & Rounds$\downarrow$ & Acc. (\%) $\uparrow$ & Rob. (\%) $\uparrow$  \\ 
\midrule
FedAvg~\cite{mcmahan2017communication} & 11.69 M & 200 & 63.58 & 0.00 \\
FedPGD~\cite{madry2018towards} & 11.69 M & 200 & 28.82 & 17.22 \\  
\midrule
VKD~\cite{hinton2015distilling} & 0.30 M & 150 & 56.78 & 3.20 \\
AKD~\cite{goldblum2020adversarially} & 0.30 M & 150 & 45.64 & 15.82  \\
PM-AFL (Ours) & 0.30 M & 150 & 45.76 & 18.74  \\
PM-AFL++ (Ours) & 0.30 M & 150 & 47.88 & 20.22 \\
\bottomrule
\end{tabular}}
\end{table}

The main contributions of this paper are summarized as follows:
\begin{itemize}
\item[$\bullet$] We focus on adversarial attacks and non-IID challenges in AFL, recognizing that training robust federated models from scratch is both computationally intensive and communication-heavy. Furthermore, we observe that neither VKD nor AKD alone is sufficient to effectively inherit both accuracy and robustness from the teacher model.
\item[$\bullet$] 
We propose the PM-AFL++ training paradigm that leverages a unified mixture KD framework to enable effective knowledge transfer between the teacher model and local models. Moreover, we introduce a global alignment term to encourage local updates to be close to global updates, thus mitigating the non-IID data challenge.
\item[$\bullet$] We conduct extensive experiments on popular benchmark datasets, along with ablation studies, to validate the effectiveness of the PM-AFL-guided training paradigm and demonstrate the indispensability of each module. To the best of our knowledge, we are the first to explore the pre-trained model-empowered AFL paradigm.
\end{itemize}

The structure of this paper is as follows. Section \ref{sec:relatedwork} reviews the related work. Section~\ref{sec:preliminary} introduces the preliminaries. Section~\ref{sec:methods} outlines the methodology, and experimental results are discussed in Section~\ref{sec:experiments}. Finally, Section~\ref{sec:conclusion} provides the conclusion.

\section{Related Work}
\label{sec:relatedwork} 
\subsection{Federated Learning}
To address privacy concerns, FL is introduced to train machine learning models in a distributed environment without requiring local data sharing. The pioneering approach, FedAvg~\cite{mcmahan2017communication}, trains a global model by aggregating model updates from multiple clients under the non-IID data challenge. Since then, various existing methods have been dedicated to further improving the performance of FedAvg from different perspectives. To mitigate the bias between local models and the global model, a mainstream line of research has concentrated on regularizing the local training process by aligning local updates with global ones. For instance, FedProx~\cite{li2020federated} introduces additional regularization terms in the local training objective, ensuring that the local model parameters do not deviate significantly from the global ones. MOON~\cite{li2021model} employs a model-contrastive approach that aligns the local model with the global model through contrastive learning, ensuring that the representations learned by the local model remain close to the global model while diverging from its previous versions. FedAvgM~\cite{hsu2019measuring} introduces momentum into the local update process, making the local training process more stable and helping to accelerate convergence. FedPer~\cite{arivazhagan2019federated} maintains shared base layers collaboratively learned across all clients while introducing personalized layers for each client, allowing adaptation to local data while preserving global knowledge. Another notable research direction leverages prototypes to improve both communication efficiency and overall performance. One notable approach in this line of work is FedProto~\cite{tan2022fedproto}, which utilizes global prototypes to guide local training and suggests transmitting prototypes instead of model parameters to enhance communication efficiency. Building on this, MP-FedCL~\cite{qiao2023mp,qiao2023framework} introduces a multi-prototype strategy to address the limitations of using a single prototype in capturing intra-class variations, enhancing the performance of the model. Furthermore, FedCCL~\cite{qiao2024fedccl} extends the multi-prototype concept to both local and global levels and proposes a parameter-free, FINCH~\cite{sarfraz2019efficient} clustering-based approach to derive local and global clustered prototypes that guide local training. Other efforts, such as FedLC~\cite{zhang2022federated} and FedCSD~\cite{yan2023rethinking}, focus on adjusting the alignment between local and global predictions from a logit perspective, while FedGen~\cite{zhu2021data} and DFRD~\cite{wang2024dfrd} utilize data-free knowledge distillation techniques. However, a major limitation of these methods is that they are designed for traditional FL scenarios, where adversarial attacks are not taken into account, leading to FL models that lack robustness against such attacks.

\subsection{Knowledge Distillation}
Knowledge distillation (KD)~\cite{hinton2015distilling} is a model compression technique that enables a smaller student model to achieve near-teacher performance by transferring compact knowledge from a high-performance teacher model, even with limited computing resources. The pioneering work~\cite{hinton2015distilling} utilizes response-based knowledge~\cite{sun2024logit} such as logit output as the information carrier for the distillation process. Additionally, researchers have explored various types of knowledge for intermediate-level guidance to better leverage additional supervision from the teacher model, including feature-based~\cite{guan2020differentiable,yu2023data} and relation-based distillation~\cite{passalis2020probabilistic,xie2024pairwise}. In the context of FL, studies~\cite{lin2020ensemble,lu2024data} have applied KD by treating the ensemble of local models as the teacher and the global model as the student, with the global model trained to match the averaged outputs of the local models. Another approach~\cite{qiao2023knowledge,qiao2023prototype,tan2022federated} treats the output, such as features from the global model, as pseudo-ground truth, encouraging the local features to align with those of the global model. Recently, researchers have also focused on improving the robustness of federated models by proposing adversarial distillation~\cite{fang2019data,goldblum2020adversarially}. This approach enhances model robustness by incorporating adversarial examples, rather than clean examples, into the distillation process. For instance, FedAdv~\cite{qiao2024knowledge} takes the first step toward prototype-based adversarial federated distillation by aligning local adversarial representations with global clean prototypes, thereby enhancing the robustness of the global model against both non-IID data and adversarial attacks. Building on this, FatCC~\cite{qiao2024logit} further extends this approach by incorporating a contrastive learning framework, where local adversarial features are encouraged to align with the corresponding global clean features while being pushed away from features of different classes. In addition, DBFAT~\cite{zhang2023delving} proposes aligning the adversarial logits of each local model with the clean logits of the global model, further enhancing the global model's adversarial robustness. However, these methods require training the model from scratch, which is computationally and communicatively demanding, particularly for large-scale models. In contrast, PM-AFL++ leverages a well-generalized, robust teacher model to transfer both accuracy and robustness to the target models, making it both communication-efficient and computationally efficient.

\subsection{Adversarial Attack and Defense}
\label{subsec:adv_attack}
Deep neural network models have been found to be vulnerable to adversarial examples (AEs), which are imperceptible to human vision~\cite{szegedy2013intriguing}. This vulnerability, first identified by~\cite{szegedy2013intriguing}, raises significant security concerns when deploying these models in real-world applications, such as autonomous vehicles~\cite{wang2023does} and security protocol-related systems~\cite{carlini2017adversarial}. Typically, adversarial attacks can be classified into white-box and black-box attacks, depending on the attacker's level of access to the model's internal information~\cite{qiao2023robustness,qiao2024federated}. In white-box attacks, the attacker has full access to the model's details, while in black-box attacks, the attacker does not have access to such information. The fast gradient sign method (FGSM)~\cite{goodfellow2014explaining} is a single-step technique for generating AEs. In contrast, projected gradient descent (PGD)~\cite{madry2018towards} and basic iterative method (BIM)~\cite{kurakin2018adversarial} are iterative extensions of FGSM that use multiple steps to craft AEs. In addition, several more advanced attack algorithms have been developed, such as the Square attack~\cite{andriushchenko2020square}, Carlini and Wagner (C\&W) attack~\cite{carlini2017towards}, and AutoAttack (AA)~\cite{croce2020reliable}. Another line of work focuses on finding a single universal attack perturbation (UAP)~\cite{moosavi2017universal} that can cause the model to misclassify all images. To defend against adversarial attacks, adversarial training (AT) is widely regarded as one of the most effective strategies~\cite{zhang2019limitations}. Recently, several studies~\cite{zizzo2020fat,qiao2024noms_fedalc,hong2023federated,chen2022calfat} have successfully applied AT in FL to develop a robust global model. FAT~\cite{zizzo2020fat} is the pioneering work that integrates the AT strategy into FL to defend against adversarial attacks. Subsequently, ~\cite{hong2023federated} proposes performing AT on resource-rich devices and sharing the resulting robustness with resource-limited devices. In addition, ~\cite{qiao2024noms_fedalc,chen2022calfat} introduce a logit calibration strategy during local AT, dynamically adjusting logit values based on class occurrence to enhance adversarial robustness. Besides, ~\cite{qiao2024logit,zhang2023delving} propose aligning local adversarial signals, such as features and logits, with their corresponding global clean counterparts to improve robustness. Orthogonal to these works, this paper explores a pre-trained model-empowered federated adversarial learning paradigm, aiming to enhance model robustness while ensuring communication efficiency.

\begin{figure*}[t]
\centering
% \raggedleft
\includegraphics[width=0.95\textwidth]{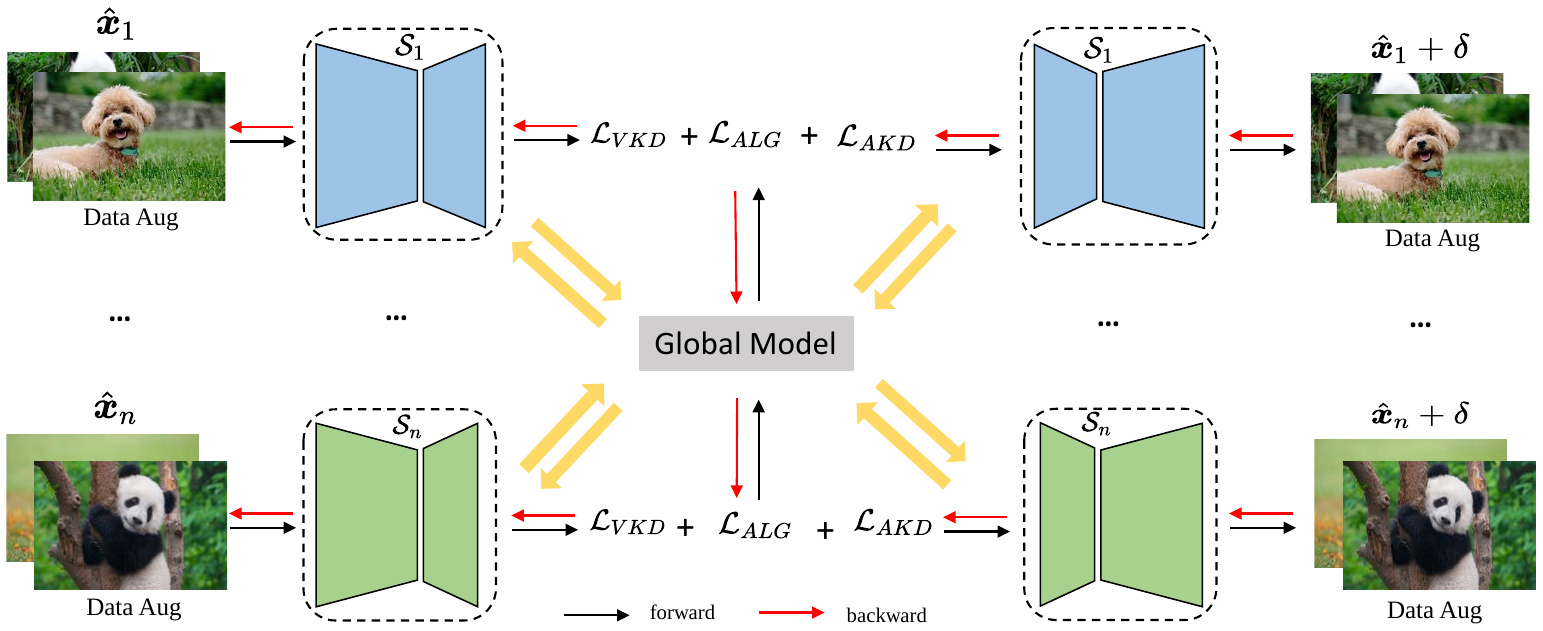}
\caption{Illustration of the proposed PM-AFL++ framework. We propose vanilla mixture knowledge distillation ($\mathcal{L}_{VKD}$) in Section~\ref{subsec:vanilla_MKD}, while adversarial mixture knowledge distillation ($\mathcal{L}_{AKD}$) is presented in Section~\ref{subsec:adversarial_MKD}. In addition, the alignment between local and global models ($\mathcal{L}_{ALG}$) is discussed in Section~\ref{subsec:alignment_local_global}.}
\label{fig:system_model}
\end{figure*}

\section{Preliminaries}
\label{sec:preliminary}
\subsection{Standard Federated Learning}
Following the standard FL setting~\cite{mcmahan2017communication,yurochkin2019bayesian}, we assume a system with $N$ clients, each holding a private dataset $\mathcal{D}_i = \{\boldsymbol{x}_i, y_i\}$ of size $D_i$. In a non-IID scenario, the label distributions across clients follow a Dirichlet distribution~\cite{balakrishnan2014continuous}, leading to varying marginal distributions $P(y)$ among clients while maintaining a consistent conditional distribution, i.e., $P_i(y | x) = P_j(y | x)$ for all clients $i$ and $j$. In addition, all clients adopt the same model architecture, with an edge server managing the collaborative training process. Each client also has access to a well-generalized, robust teacher model. Under these conditions, the local training objective for each client can be formulated as follows:
\begin{equation} \label{eq_local_loss}
   {\mathcal{L}_i} (\omega_i) = - \frac {1}{D_i} \sum_{i \in \mathcal{D}_i}\sum_{j=1}^{C}\mathbbm{1}_{y = j}\log \frac{e^{z_{i,j}}}{\sum_{j=1}^{C}e^{z_{i,j}}},
\end{equation}
where $z$ represents the model output, which is further aligned with the output of the teacher model. Here, $\mathbbm{1}(\cdot)$ denotes the indicator function, $\omega_i$ represents the model parameters, and $C$ is the total number of classes. 

Next, each client updates its local model parameters using stochastic gradient descent to minimize its local objective:
\begin{equation} \label{sgd}
\omega_{t+1}  =  \omega_{t} - \eta \nabla \mathcal{L}_i(\omega_{t}; \boldsymbol{x}_i, y_i),
\end{equation}
where $\nabla \mathcal{L}_i(\omega_{t}; \boldsymbol{x}_i, y_i)$ denotes the gradient of the loss function for client $i$ in the current round, $\omega_{t+1}$ represents the updated model parameters for the next round, and $\eta$ is the learning rate.

Finally, the global objective is then to aggregate the local losses across all distributed clients as follows:
\begin{equation} \label{eq_global_loss}
    \mathcal{L} (\omega) = \sum_{i \in [N]} \frac {D_i}{\sum_{i \in [N]}D_i} {\mathcal{L}_i} (\omega_i),
\end{equation}
where $\omega$ denotes the global model parameters, and $[N]$ represents the set of distributed clients, defined as $[N] = \{1,...,N\}$. The overall objective is to enhance the robustness of the global model by leveraging knowledge distillation (KD) from a well-trained teacher model to local models during training, ultimately enhancing the global model after each communication round.

\subsection{Adversarial Attacks Meet Federated Learning}
\label{subsec:notation_adv_attack}
Adversarial attacks can easily mislead a model by introducing carefully crafted, imperceptible perturbations, resulting in incorrect predictions~\cite{szegedy2013intriguing}. For any given client, the classification layer of the model is represented as $\phi_i(\boldsymbol{x}_i): \mathbb{R}^{h \times w \times c}\rightarrow [C]$, mapping the input image $\boldsymbol{x}_i$ to a discrete set of labels $[C]$, where $h$, $w$, and $c$ denote the height, width, and number of channels of the image, respectively. To find a well-crafted perturbation $\delta \in \mathbb{R}^{h \times w \times c}$ that causes $\phi(\boldsymbol{x}_i + \delta) \neq \phi(\boldsymbol{x}_i)$, we use the PGD attack to iteratively generate the AEs as follows:
\begin{equation}
\label{eq:pgd}
\boldsymbol x_i^{t+1}=\Pi_{\boldsymbol{x_i} + \delta}\left(\boldsymbol x_i^t+\alpha\sign(\nabla_{\boldsymbol x_{i}}  \mathcal{L}_i(\omega_i; \boldsymbol{x}_i^t, y_i) \right),
\end{equation}
where $\alpha$ denotes the step size, $\boldsymbol x_i^t$ represents the AE generated at the $t$-th step, $\Pi_{\boldsymbol x_i + \delta}(\cdot)$ projects the perturbed input into the feasible region $\boldsymbol{x_i} + \delta$, and $\sign(\cdot)$ denotes the sign function. To ensure that the perturbation $\delta$ remains imperceptible to human vision, it is typically constrained by an upper bound $\epsilon$. Consequently, in each iteration of the PGD attack, the optimal perturbation $\delta^{\ast}$ is obtained by maximizing the local objective in (\ref{eq_local_loss}), as follows:
\begin{equation} \label{delta_optimi}
   \delta^{\ast} = \mathop {\arg\max}_{||\delta||_\infty \leq \epsilon} \mathcal{L}_i(\omega_i; \boldsymbol{x}_i + \delta, y_i),
\end{equation}
where $\delta^{\ast}$ denotes the perturbation obtained after the predefined number of iterations of the PGD attack algorithm. Upon completion of these iterations, the AEs $\boldsymbol{x}_i^{adv}$ can be expressed as follows:
\begin{equation} \label{AE_Gen}
   \boldsymbol{x}_i^{adv} = \boldsymbol{x}_i + \delta^{\ast}.
\end{equation}

To defend against such attacks, a common approach in existing works~\cite{chen2022calfat,qiao2024logit,zhang2023delving,hong2023federated} is to incorporate adversarial training (AT) into the local training phase of FL, as AT is a well-established and widely recognized defense method~\cite{athalye2018robustness}. Specifically, the AEs generated in (\ref{AE_Gen}) are used as new inputs for each local training process. By minimizing the loss with these AEs, each local model is expected to improve its robustness against such attacks. The final objective can then be formulated as follows:
\begin{equation} \label{AT_federated}
   \mathop {\min}_{\omega} \mathbb E_{(\boldsymbol{x}_i^{adv}, y_i) \sim \mathcal{D}_i}  \mathcal{L}_i(\omega_i; \boldsymbol{x}_i^{adv}, y_i).
\end{equation}

However, as demonstrated in Table~\ref{tab:motivation}, training a robust model from scratch using pure AT is resource-intensive. In contrast, KD-based approaches yield promising results with fewer resources and improved performance. This motivates us to explore how KD can be leveraged to develop a robust and generalizable global model with lower resource requirements.

\section{Towards Communication-Efficient Adversarial Federated Learning}
\label{sec:methods}
Knowledge distillation~\cite{hinton2015distilling} is a natural choice for improving the performance of smaller student models. It transfers the teacher model's knowledge, including its accuracy and generalization capabilities, to resource-constrained student models. This enables the student models to approach the performance of the teacher model without requiring the same computational resources as training a large model from scratch~\cite{qiao2024knowledge,hinton2015distilling}. In this paper, we adopt this approach and further explore the use of vanilla and adversarial mixture knowledge distillation to transfer both the accuracy and robustness of a teacher model to local models within a unified framework.

\subsection{Vanilla Mixture Knowledge Distillation}
\label{subsec:vanilla_MKD}
In this paper, vanilla mixture knowledge distillation refers to the process of transferring knowledge from the teacher model to student models using both clean samples and augmented clean samples as inputs. We assume that this distillation process is performed on an arbitrary client, and for simplicity, we omit the client subscript. Specifically, given two distinct clean images, $\boldsymbol{x}_i$ and $\boldsymbol{x}_j$, we mix them using a combination factor $\lambda$~\cite{zhang2017mixup}, which controls the mixing ratio, as shown below:
\begin{equation} \label{clean_mixup}
\hat{\boldsymbol{x}}_{ij} = \lambda \boldsymbol{x}_i + (1 - \lambda) \boldsymbol{x}_j
\end{equation}
where $\hat{\boldsymbol{x}}_{ij}$ denotes an augmented image and $\lambda$ is sampled from $\text{Beta}(\beta, \beta)$ with $\beta \in (0, +\infty)$.

Subsequently, we feed the clean images and their mixed version into the teacher and student models, respectively. For the teacher model, the outputs corresponding to the clean images and their mixed version are defined as follows:
\begin{equation} \label{eq:mixup_teacher}
\begin{aligned}
z_{ij}^t &= \lambda \mathcal{T}(\boldsymbol{x}_i) + (1 - \lambda) \mathcal{T}(\boldsymbol{x}_j), \\
\hat{z}_{ij}^t &= \mathcal{T}(\hat{\boldsymbol{x}}_{ij}),
\end{aligned}
\end{equation}
where $z_{ij}^t$ represents the teacher model's linearly interpolated class probabilities based on the inputs $\boldsymbol{x}_i$ and $\boldsymbol{x}_j$, and $\hat{z}_{ij}^t$ denotes the class probabilities for the augmented image $\hat{\boldsymbol{x}}_{ij}$. Here, $\mathcal{T}(\boldsymbol{x})$ represents the output of the teacher model with clean sample $\boldsymbol{x}$ as input. Similarly, the outputs of the student model are defined as follows:
\begin{equation} \label{clean_stu_mixup}
\begin{aligned} 
z_{ij}^s &= \lambda \mathcal{S}(\boldsymbol{x}_i) + (1 - \lambda) \mathcal{S}(\boldsymbol{x}_j), \\
\hat{z}_{ij}^s &= \mathcal{S}(\hat{\boldsymbol{x}}_{ij}),
\end{aligned}
\end{equation}
where $z_{ij}^s$ represents the local model's linearly interpolated class probabilities based on the inputs $\boldsymbol{x}_i$ and $\boldsymbol{x}_j$, and $\hat{z}_{ij}^s$ denotes the class probabilities of the augmented image $\hat{\boldsymbol{x}}_{ij}$. Here, $\mathcal{S}(\boldsymbol{x})$ represents the output of the student model with clean sample $\boldsymbol{x}$ as input.

To encourage the teacher model to provide the student model with more diverse distillation targets, we propose distilling knowledge between pairs of clean samples from the teacher model and their corresponding student outputs. Similarly, we also perform distillation using the mixed version of the input for the teacher model, transferring knowledge to the student model with the corresponding mixed version as input. Therefore, the vanilla knowledge distillation (VKD) process can be defined as follows:
\begin{equation} 
\label{mixture_vanilla_KD}
   \mathcal{L}_{VKD} = KL(z_{ij}^t, z_{ij}^s) + KL(\hat{z}_{ij}^t, \hat{z}_{ij}^s),
\end{equation}
where $\mathcal{L}_{VKD}$ denotes the vanilla distillation process and $KL(\cdot)$ represents the Kullback-Leibler divergence~\cite{ji2020kullback} loss. Note that in this distillation process, the local model is optimized, while the teacher model's parameters are fixed.

\subsection{Adversarial Mixture Knowledge Distillation} 
\label{subsec:adversarial_MKD}
Similar to vanilla mixture knowledge distillation, we define adversarial mixture knowledge distillation as the process of transferring knowledge from the teacher model to student models using both adversarial samples and augmented adversarial samples as inputs. Specifically, given two distinct clean images, $\boldsymbol{x}_i$ and $\boldsymbol{x}_j$, we first generate their corresponding adversarial samples using (\ref{eq:pgd}) with the constraints in (\ref{delta_optimi}). The generated adversarial samples are denoted as $\boldsymbol{x}_i^{adv}$ and $\boldsymbol{x}_j^{adv}$, respectively. We then mix these adversarial samples using a combination factor $\lambda$~\cite{zhang2017mixup}, which controls the mixing ratio, as shown below:
\begin{equation} \label{ae_mixup}
\hat{\boldsymbol{x}}_{ij}^{adv} = \lambda \boldsymbol{x}_i^{adv} + (1 - \lambda) \boldsymbol{x}_j^{adv},
\end{equation}
where $\hat{\boldsymbol{x}}_{ij}^{adv}$ denotes the augmented AEs, and $\lambda$ is sampled from $\text{Beta}(\beta, \beta)$ with $\beta \in (0, +\infty)$.

Subsequently, we feed the generated adversarial samples and their mixed version into the student models. The outputs corresponding to the adversarial samples and their mixed version are defined as follows:
\begin{equation} \label{adv_stu_mixup}
\begin{aligned}
z_{ij}^{s,adv} &= \lambda \mathcal{S}(\boldsymbol{x}_i^{adv}) + (1 - \lambda) \mathcal{S}(\boldsymbol{x}_j^{adv}), \\
\hat{z}_{ij}^{s,adv} &= \mathcal{S}(\hat{\boldsymbol{x}}_{ij}^{adv}),
\end{aligned}
\end{equation}
where $z_{ij}^{s,adv}$ represents the local model's linearly interpolated class probabilities based on the inputs $\boldsymbol{x}_i^{adv}$ and $\boldsymbol{x}_j^{adv}$, and $\hat{z}_{ij}^{s,adv}$ denotes the class probabilities for the augmented image $\hat{\boldsymbol{x}}_{ij}^{adv}$. Here, $\mathcal{S}(\boldsymbol{x}^{adv})$ represents the output of the student model with the adversarial sample as input.

Inspired by~\cite{kannan2018adversarial, engstrom2018evaluating, wu2023adaptive}, which suggest that aligning adversarial logits with clean logits can enhance model robustness, we propose aligning the adversarial sample outputs of the student model with the corresponding clean outputs of the teacher model. Similarly, following the vanilla mixture distillation approach, we also encourage alignment between the mixed adversarial sample outputs of the student model and the corresponding mixed clean outputs of the teacher model. Therefore, we define the adversarial knowledge distillation (AKD) process as follows:
\begin{equation} 
\label{mixture_adversarial_KD}
   \mathcal{L}_{AKD} = KL(z_{ij}^t, z_{ij}^{s,adv}) + KL(\hat{z}_{ij}^t, \hat{z}_{ij}^{s,adv}),
\end{equation}
where $\mathcal{L}_{AKD}$ denotes the adversarial distillation process and $KL(\cdot)$ represents the Kullback-Leibler divergence~\cite{ji2020kullback} loss. Again, in this distillation process, the local model is optimized, while the teacher model's parameters are fixed.

\begin{algorithm}[t] 
\centering
    \caption{PM-AFL++} 
    \label{alg:PM-AFL} 
    \begin{algorithmic}[1]
        \REQUIRE ~~ \\
        Private dataset $\mathcal{D}_i$ for each client, initialized model $\omega$, teacher model $\mathcal{T}$, number of clients $N$, global rounds $T$.
        \ENSURE ~~ \\
        Robust global model.
        \FOR{ $t$ = 1, 2, ..., $T$} 
            \FOR{ $i$ = 0, 1,..., $N$ \textbf{in parallel}}
                \STATE Send global model $\omega^t$ to local client \textit{i}
                \STATE {$\omega^t \gets \textbf{LocalUpdate}(\omega^t$)}
            \ENDFOR
            \STATE {$\mathcal{L} (\omega) \gets \sum_{i \in [N]} \frac {D_i}{\sum_{i \in [N]}D_i} {\mathcal{L}_i} (\omega_i)$ by (\ref{eq_global_loss})}
        \ENDFOR \\
        \textbf {LocalUpdate($\omega^t$)}
        \FOR{ each local epoch }
        \FOR{ each batch ($\boldsymbol{x}_i$; $y_i$) of $\mathcal{D}_i$}
        \STATE \textcolor{gray}{\text{/* Adversarial examples generation */}}
        \STATE {$\boldsymbol{x}_i^{adv} \gets \boldsymbol{x}_i + \delta^{\ast}$} by (\ref{AE_Gen})
        \STATE {\textcolor{gray}{\text{/* Clean examples augmentation */}}}
        \STATE {$\tilde{\boldsymbol{x}}_{ij} \gets \lambda \boldsymbol{x}_i + (1 - \lambda) \boldsymbol{x}_j$ via (\ref{clean_mixup})}
        \STATE {\textcolor{gray}{\text{/* Adversarial examples augmentation */}}}
        \STATE{$\tilde{\boldsymbol{x}}_{ij}^{adv} \gets \lambda \boldsymbol{x}_i^{adv} + (1 - \lambda) \boldsymbol{x}_j^{adv}$ via (\ref{ae_mixup})}
        \STATE {\textcolor{gray}{\text{/* Vanilla mixture knowledge distillation */}}}
        \STATE {$ \mathcal{L}_{VKD} \gets KL(z_{ij}^t, z_{ij}^s) + KL(\hat{z}_{ij}^t, \hat{z}_{ij}^s)$} via (\ref{mixture_vanilla_KD})
        \STATE {\textcolor{gray}{\text{/* Adversarial mixture knowledge distillation */}}}
        \STATE {$\mathcal{L}_{AKD} \gets KL(z_{ij}^t, z_{ij}^{s,adv}) + KL(\hat{z}_{ij}^t, \hat{z}_{ij}^{s,adv})$} via (\ref{mixture_adversarial_KD})
        \STATE {\textcolor{gray}{\text{/* Consistency regularization */}}}
        \STATE {$\mathcal{L}_{ALG} \gets \Vert z_s^{adv}- z_g)\|_2^2$} via (\ref{reg_local_global})
        \STATE {\textcolor{gray}{\text{/* Overall local objective for each client */}}}
        \STATE{$\mathcal{L} \gets \alpha \mathcal{L}_{VKD} + (1-\alpha) \mathcal{L}_{AKD} + \mathcal{L}_{ALG}$ via (\ref{overall_KD})}
    \ENDFOR
    \ENDFOR
        \RETURN $\omega_i^t$
    \end{algorithmic}
\end{algorithm}

\subsection{Alignment Between Local and Global}
\label{subsec:alignment_local_global}
However, due to the non-IID distribution across clients, the update directions of local models may deviate from that of the global model, potentially causing misalignment. To address this, we introduce a consistency regularization term that encourages each local adversarial representation to align with the corresponding global clean representations. During each global communication round, for an arbitrary client $i$, the local adversarial representation $z_s^{adv}$ is obtained using the local student model $\mathcal{S}(\boldsymbol{x}_i^{adv})$ with adversarial sample $\boldsymbol{x}_i^{adv}$ as input, while $z_g$ is derived from the global model using clean samples $\boldsymbol{x}_i$ as input. The local adversarial representations are then aligned with the global clean representations by minimizing the mean squared error. Therefore, the alignment between local and global (ALG) can be defined as follows:
\begin{equation} 
\label{reg_local_global}
\mathcal{L}_{ALG} = \Vert z_s^{adv}- z_g)\|_2^2,
\end{equation}
where $\Vert \cdot \Vert_2^2$ denotes the squared $\ell_2$ distance used to measure the difference between the local adversarial features and the global clean ones.

\begin{table*}[!htbp]
\centering
\caption{Comparison of different methods on benchmark datasets. The best results are in \textbf{bold} and second with \underline{underline}. PM-AFL and PM-AFL++ outperform the baselines in most cases, with PM-AFL++ achieving higher accuracy (\%) and robustness (\%) while requiring significantly fewer communication parameters.} 
\vspace{-3px}
\label{tab:performance_compar}
\setlength{\abovecaptionskip}{0.1cm}
\setlength{\belowcaptionskip}{-0.07cm}
\resizebox{\textwidth}{!}{
\begin{tabular}{llcccccccccc}
\toprule
\multirow{2}*{\textbf{Dataset}} & \multirow{2}*{Method} & \multirow{2}*{Clean Acc.} & \multicolumn{7}{c} {Robust Acc.} & \multirow{2}*{\begin{tabular}[c]{@{}c@{}}\# of Comm \\ Rounds \end{tabular}} & \multirow{2}*{\begin{tabular}[c]{@{}c@{}}\# of Comm \\ Params ($\times 10^3$) \end{tabular}} \\
\cmidrule{4-10}
& & & FGSM & BIM & PGD-40 & PGD-100 & Square & AA & \textbf{Avg} & \\
\midrule
\multirow{10}*{\textbf{MNIST}} & FedAvg~\cite{mcmahan2017communication} & 90.54 & 37.64 & 0.68 & 0.00 & 0.00 & 0.16 & 0.00 & 6.41 & 160 & 3,217 \\
& MixFAT~\cite{zizzo2020fat} & 91.30 & 52.92 & 48.78 & 15.24 & 6.32 & 0.70 & 0.04 & 20.66 & 160 & 3,217 \\
& FedPGD~\cite{madry2018towards} & 91.54 & 51.74 & 52.46 & 18.06 & 7.72 & 0.40 & 0.06 & 21.74 & 160 & 3,217 \\
& FedALP~\cite{kannan2018adversarial} & 94.06 & 64.62 & 60.02 & 29.24 & 12.18 & 1.26 & 0.76 & 28.01 & 180 & 3,217 \\
& FedMART~\cite{wang2019improving} & 93.74 & 61.24 & 47.52 & 16.76 & 7.20 & 0.76 & 0.26 & 22.29 & 160 & 3,217 \\
& FedTRADES~\cite{zhang2019theoretically} & 94.32 & \textbf{66.26} & 51.80 & 17.92 & 4.46 & 0.46 & 0.04 & 23.49 & 160 & 3,217 \\
& CalFAT~\cite{chen2022calfat} & 93.60 & 64.48 & 45.38 & 14.78 & 1.68 & 0.20 & 0.06 & 21.09 & 180 & 3,217 \\
& DBFAT~\cite{zhang2023delving} & 93.58 & \underline{66.14} & 61.70 & \underline{37.62} & 16.82 & 0.44 & 0.34 & 30.51 & 180 & 3,217 \\
& PM-AFL (Ours) & \underline{94.53} & 57.89 & \underline{72.15} & 33.41 & \underline{18.72} & \underline{17.84} & \underline{13.66} & \underline{35.61} & \textbf{100} & \textbf{44} \\
& PM-AFL++ (Ours) & \textbf{94.68} & 63.96 & \textbf{77.50} & \textbf{43.10} & \textbf{29.92} & \textbf{27.52} & \textbf{23.62} & \textbf{44.27} & \textbf{100} & \textbf{44} \\
\midrule
\multirow{10}*{\textbf{CIFAR-10}} & FedAvg~\cite{mcmahan2017communication} & 63.58 & 3.22 & 0.00 & 0.00 & 0.00 & 0.36 & 0.00 & 0.59 & 200 &  11,690 \\
& MixFAT~\cite{zizzo2020fat} & 38.94 & 22.68 & 21.24 & 21.32 & 21.22 & 20.48 & 18.08 & 20.83 & 250 & 11,690\\
& FedPGD~\cite{madry2018towards} & 28.82 & 19.82 & 19.50 & 19.48 & 19.42 & 18.36 & 17.22 & 18.96 & 200 &  11,690 \\
& FedALP~\cite{kannan2018adversarial} & 31.54 & 21.18 & 20.15 & 20.10 & 20.08 & 18.70 & 16.86 & 19.51 & 200 & 11,690 \\
& FedMART~\cite{wang2019improving} & 35.34 & 22.67 & 20.64 & 20.15 & 19.93 & 19.13 & 17.82 & 20.05 & 200 & 11,690 \\
& FedTRADES~\cite{zhang2019theoretically} & 36.00 & 21.06 & 19.62 & 19.64 & 19.54 & 19.80 & 17.32 & 19.49 & 230 & 11,690 \\
& CalFAT~\cite{chen2022calfat} & 32.24 & 20.98 & 19.66 & 19.68 & 19.60 & 18.72 & 16.98 & 19.27 & 200 & 11,690 \\
& DBFAT~\cite{zhang2023delving} & 30.82 & 18.32 & 18.00 & 17.92 & 17.90 & 17.42 & 16.64 & 17.70 & 200 & 11,690 \\
& PM-AFL (Ours) & \underline{45.76} & \underline{24.46} & \underline{23.96} & \underline{22.96} & \underline{22.94} & \underline{21.26} & \underline{18.74} & \underline{22.38} & \textbf{150} & \textbf{320} \\
& PM-AFL++ (Ours) & \textbf{47.88} & \textbf{26.80} & \textbf{24.62} & \textbf{24.68} & \textbf{24.66} & \textbf{23.20} & \textbf{20.22} & \textbf{24.03} & \textbf{150} & \textbf{320} \\
\midrule
\multirow{10}*{\textbf{CIFAR-100}} & FedAvg~\cite{mcmahan2017communication} & 50.81 & 0.00 & 0.00 & 0.00 & 0.00 & 0.60 & 0.00 & 0.10 & 200 & 11,690 \\
& MixFAT~\cite{zizzo2020fat} & 46.26 & 23.40 & 18.83 & 18.80 & 18.60 & 22.20 & 17.20 & 19.83 & 200 & 11,690  \\
& FedPGD~\cite{madry2018towards} & 45.60 & 22.60 & 19.77 & 19.80 & 19.74 & 21.40 & 17.80 & 20.18 & 220 & 11,690  \\
& FedALP~\cite{kannan2018adversarial} & 47.41 & 21.80 & 20.20 & 20.40 & 20.03 & 21.07 & 18.80 & 20.38 & 200 & 11,690  \\
& FedMART~\cite{wang2019improving} & 46.62 & 22.01 & 15.99 & 16.07 & 15.66 & 19.88 & 14.12 & 17.28 & 250 & 11,690 \\
& FedTRADES~\cite{zhang2019theoretically} & 48.29 & 21.84 & 17.16 & 16.86 & 16.54 & 20.44 & 15.96 & 18.13 & 200 & 11,690  \\
& CalFAT~\cite{chen2022calfat} & 49.09 & 22.11 & 18.09 & 18.02 & 17.82 & 20.40 & 17.03 & 18.91 & 200 & 11,690 \\
& DBFAT~\cite{zhang2023delving} & 48.58 & 21.80 & 19.44 & 19.22 & 18.86 & 20.51 & 18.23 & 19.67 & 200 & 11.690  \\
& PM-AFL (Ours) & \underline{54.41} & \underline{31.60} & \underline{28.90} & \underline{28.88} & \underline{28.71} & \underline{26.20} & \underline{19.90} &  \underline{27.36}  & \textbf{150} & \textbf{504} \\
& PM-AFL++ (Ours) & \textbf{57.40} & \textbf{33.81} & \textbf{30.22} & \textbf{30.10} & \textbf{30.09} & \textbf{27.60} & \textbf{22.08} & \textbf{28.98} & \textbf{150} & \textbf{504}  \\
\bottomrule
\end{tabular}}
\end{table*}

\subsection{Overall Objective} 
Our proposed PM-AFL++ framework is built upon three key components. First, to improve the clean accuracy of the global model, we introduce vanilla mixture distillation, which inherits the clean accuracy from the teacher model by transferring knowledge from both clean samples and their mixed counterparts to the student model. Second, to improve the adversarial robustness of the global model, we propose adversarial mixture distillation, which enhances robustness by aligning adversarial samples and their mixed counterparts with the corresponding clean outputs of the teacher model. Note that both strategies are integrated into a unified framework, with an introduced coefficient to balance the trade-off between clean accuracy and robust accuracy. Finally, to address the challenge of non-IID data among clients, we introduce an alignment term that encourages consistency between the local and global models by aligning local adversarial representations with their corresponding global clean representations. By jointly optimizing these three components, local models are expected to achieve a balance between accuracy and robustness while mitigating the risk of overfitting to their own data distributions. As a result, each client benefits from these objectives, leading to the formulation of the overall objective function as follows:
\begin{equation} \label{overall_KD}
   \mathcal{L} = \alpha \mathcal{L}_{VKD} + (1-\alpha) \mathcal{L}_{AKD} + \mathcal{L}_{ALG},
   % \vspace{-3px}
\end{equation}
where $\mathcal{L}$ represents the overall local objective, and $\alpha$ is a weighting factor that controls the trade-off between accuracy and robustness. The detailed training procedure of the proposed framework is outlined in Algorithm~\ref{alg:PM-AFL}. In each global round, clients receive the model parameters from the server (line 3) and then perform local training (lines 8 to 26). During local training, clients compute vanilla mixture knowledge distillation, adversarial mixture knowledge distillation, and consistency regularization in lines 17, 19, and 21, respectively. Based on these computations, clients update their model parameters (line 23) and send the updated parameters back to the server (line 26). The server then aggregates all the training parameters (line 6) and initiates the next global round until the required number of global rounds is completed.

\section{Experiments}
\label{sec:experiments}
\subsection{Experimental Setup}
\textbf{Datasets and Baselines.} We conduct experiments on three widely used benchmark datasets: MNIST~\cite{lecun1998gradient}, CIFAR-10~\cite{krizhevsky2009learning}, and CIFAR-100~\cite{krizhevsky2009learning}, to verify the effectiveness of the proposed PM-AFL framework, including PM-AFL++. Since the research on AFL is still in its early stages with limited established methods, we incorporate four well-established defense methods, including PGD\_AT~\cite{madry2018towards}, ALP~\cite{kannan2018adversarial}, MART~\cite{wang2019improving}, and TRADES~\cite{zhang2019theoretically} into the AFL framework, and refer to them as FedPGD, FedALP, FedMART, and FedTRADES, respectively. In addition, for a more comprehensive evaluation, we compare PM-AFL and PM-AFL++ with three other state-of-the-art federated defense methods such as MixFAT~\cite{zizzo2020fat}, CalFAT~\cite{chen2022calfat}, and DBFAT~\cite{zhang2023delving}, as well as FedAvg~\cite{mcmahan2017communication}, which denotes typical federated training without adversarial training process. To evaluate the effectiveness of our proposed method, we utilize five mainstream attack techniques such as FGSM~\cite{goodfellow2014explaining}, BIM~\cite{kurakin2018adversarial}, PGD~\cite{madry2018towards}, Square~\cite{andriushchenko2020square}, and AA~\cite{croce2020reliable}.

\textbf{Implementation Details.} 
Following~\cite{chen2022calfat,zhang2023delving}, we adopt simple CNN models as the local models for the MNIST, CIFAR-10, and CIFAR-100 tasks in both PM-AFL and PM-AFL++. For the teacher models, following~\cite{zagoruyko2016wide,rebuffi2021fixing}, we utilize the pre-trained WideResNet-28-10~\cite{zagoruyko2016wide} for the MNIST task, WideResNet-34-10~\cite{zagoruyko2016wide} for the CIFAR-10 task, and WideResNet-28-10~\cite{rebuffi2021fixing} for the CIFAR-100 task. Note that the teacher model is used locally for forward propagation only and is not sent to the server for aggregation. To further reduce computation, its predictions can also be saved locally, allowing each student model to align with them through a single forward pass. Nevertheless, since the teacher model is only used locally and can save predictions in advance, the proposed framework will not incur additional communication costs. For the baselines, we adopt MobileNet~\cite{howard2017mobilenets} for MNIST and ResNet-18~\cite{he2016deep} for both CIFAR-10 and CIFAR-100. To simulate non-IID settings, we employ the Dirichlet distribution Dir($a$)~\cite{yurochkin2019bayesian}, with $a$ set to 0.1 by default. Following~\cite{zhang2017mixup}, we set $\lambda$ to 0.2 for the MNIST, CIFAR-10, and CIFAR-100 datasets. In addition, following~\cite{goodfellow2014explaining}, we set the perturbation bound to 8/255 and the step size to 2/255 for both CIFAR-10 and CIFAR-100, while for MNIST, we set the perturbation bound to 0.3 and the step size to 0.01. Final performance is evaluated by calculating the mean of the last five communication rounds, and all experimental results are averaged over three independent runs.

\begin{figure}[t]
\centering
\includegraphics[width=0.42\textwidth]{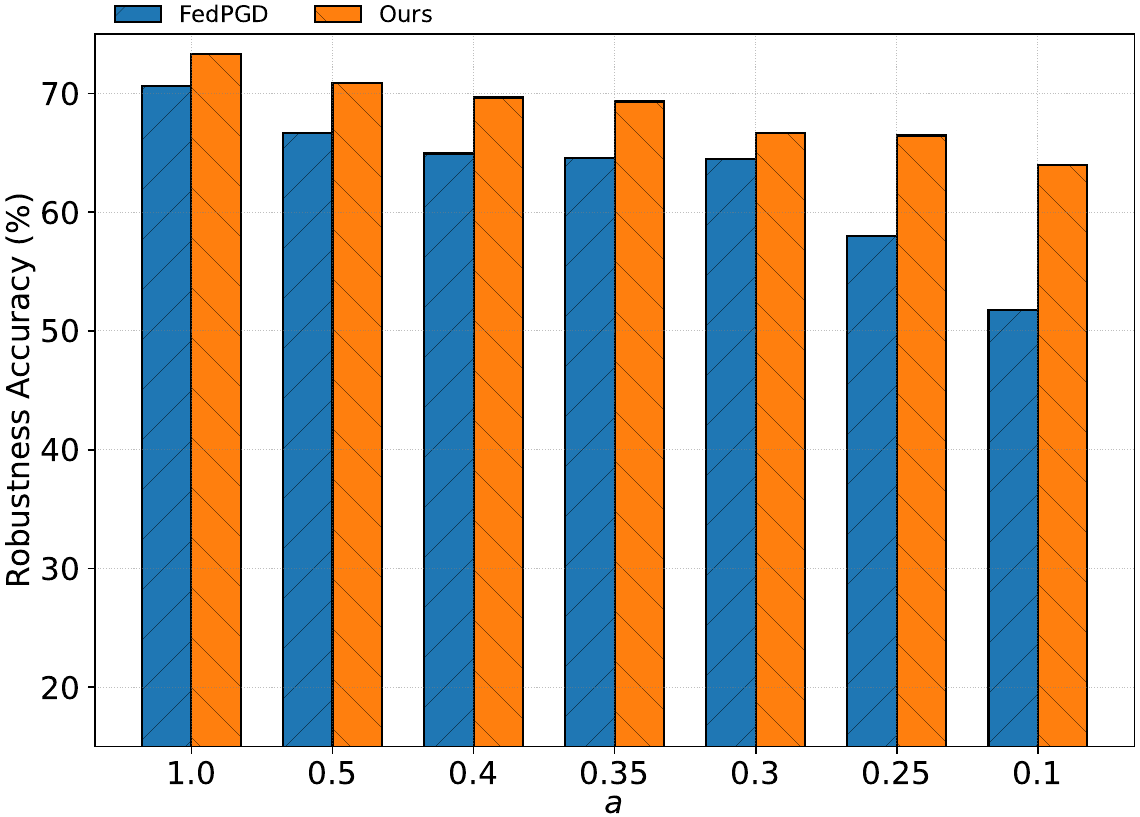}
\vspace{-3pt}
\caption{Robustness comparison of PM-AFL++ and FedPGD on MNIST under different levels of heterogeneities.}
\label{fig:fgsm_scala_comparision}
\vspace{-3pt}
\end{figure}

\begin{figure}[t]
\centering
\includegraphics[width=0.42\textwidth]{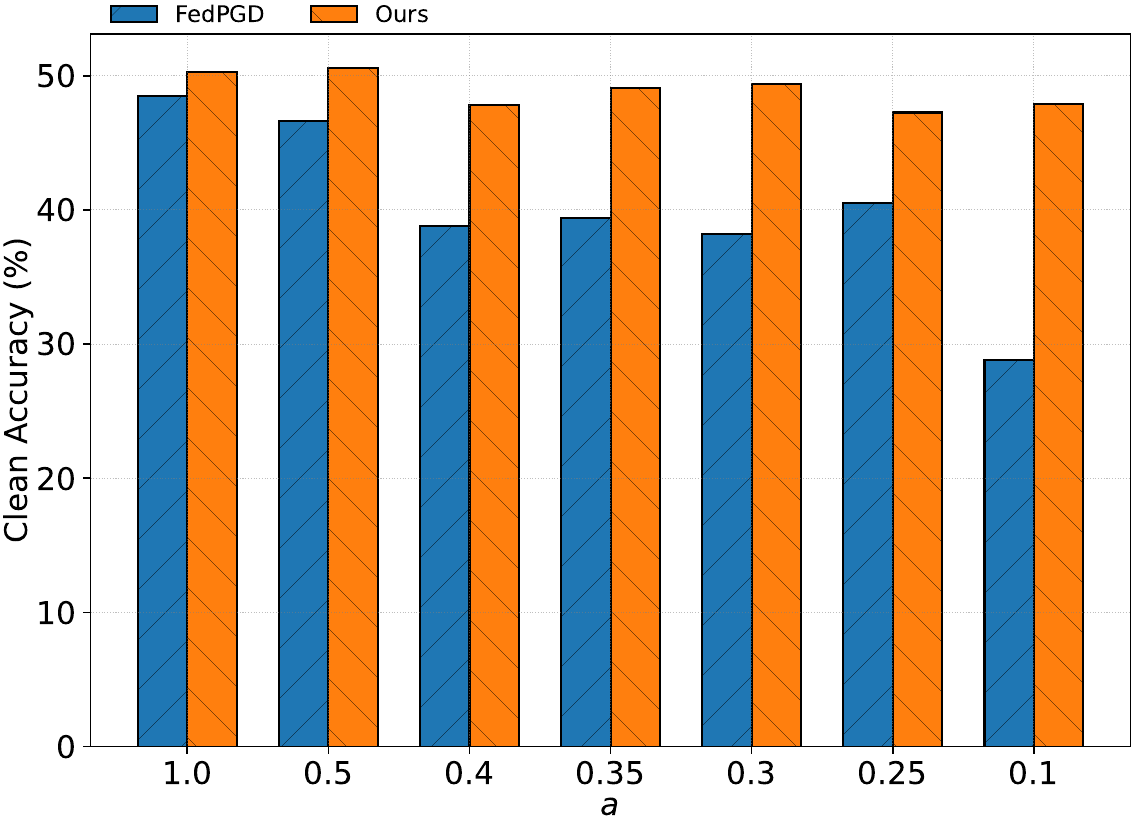}
\vspace{-3pt}
\caption{Accuracy comparison of PM-AFL++ and FedPGD on CIFAR-10 under different levels of heterogeneities.}
\label{fig:clean_scala_comparision}
\vspace{-3pt}
\end{figure}

\subsection{Performance Comparison}
\textbf{Accuracy Comparision.} We present the performance comparison in Table~\ref{tab:performance_compar}, including clean and robust accuracy. Clean accuracy is measured on unperturbed samples, while robust accuracy is evaluated using six attack metrics: FGSM, BIM, PGD-40, PGD-100, Square, and AA. Additionally, we also report the average robust accuracy across these attacks for a comprehensive assessment of model robustness. Several key observations can be drawn from the table. First, adversarial attacks pose a significant challenge to the clean accuracy of federated models. For example, in the CIFAR-10 task, the clean accuracy of FedAvg dramatically declines from 63.58\% to an average of merely 0.59\% under six different attacks. This highlights the need for defense strategies against adversarial attacks in the context of AFL. Second, while existing defense mechanisms improve adversarial robustness compared to FedAvg, their effectiveness remains limited and often comes at the expense of clean accuracy. For instance, in the CIFAR-10 task, FedPGD increases AA accuracy from 0.00\% to 17.22\%, but this comes with a significant drop in clean accuracy from 63.58\% to 28.82\%. Third, our strategies, PM-AFL and PM-AFL++, particularly PM-AFL++, improve model robustness while preserving relatively high clean accuracy compared to other federated defense methods. For example, in the MNIST tasks, PM-AFL++ achieves a clean accuracy of 94.68\% and also performs well in robust accuracy, particularly under BIM attacks, with a score of 77.50\%. Similarly, in CIFAR-10 tasks, PM-AFL++ attains a clean accuracy of 47.88\% while significantly improving robust accuracy against various attacks, such as FGSM and BIM. These results underscore that PM-AFL, particularly PM-AFL++, provides a significant advantage in improving both clean accuracy and robustness against adversarial attacks and non-IID data challenges.

\begin{table}[t]
\centering
\caption{Comparison of different configurations. CA denotes clean accuracy, while RA represents robust accuracy. The optimal trade-off is in \textbf{bold}.}
\label{tab:ablation_comparison}
\vspace{-5px}
\resizebox{0.95\linewidth}{!}{
\begin{tabular}{l|ccc|cc}
\toprule
Dataset & $\mathcal{L}_{VKD}$ & $\mathcal{L}_{AKD}$ & $\mathcal{L}_{ALG}$ & CA (\%) & RA (\%) \\
\midrule
\multirow{4}*{MNIST}
& \xmark & \xmark & \xmark & 91.54 & 21.74 \\
& \cmark & \xmark & \xmark & 94.34 & 2.59 \\
& \cmark & \cmark & \xmark & 94.40 & 36.41 \\
& \cmark & \cmark & \cmark & \textbf{94.68} & \textbf{44.27} \\
\midrule
\multirow{4}*{CIFAR-10}
& \xmark & \xmark & \xmark & 28.82 & 18.96 \\
& \cmark & \xmark & \xmark & 57.32 & 8.83 \\
& \cmark & \cmark & \xmark & 47.12 & 23.88 \\
& \cmark & \cmark & \cmark & \textbf{47.88} & \textbf{24.03} \\
\midrule
\multirow{4}*{CIFAR-100}
& \xmark & \xmark & \xmark & 45.60 & 20.18 \\
& \cmark & \xmark & \xmark & 69.84 & 4.24 \\
& \cmark & \cmark & \xmark & 47.48 & 23.87 \\
& \cmark & \cmark & \cmark & \textbf{57.40} & \textbf{28.98} \\
\bottomrule
\end{tabular}}
\vspace{-5px}
\end{table}

\begin{figure*}[t]
\centering
    \begin{minipage}[b]{0.32\textwidth}
        \centering
         \includegraphics[width=\textwidth]{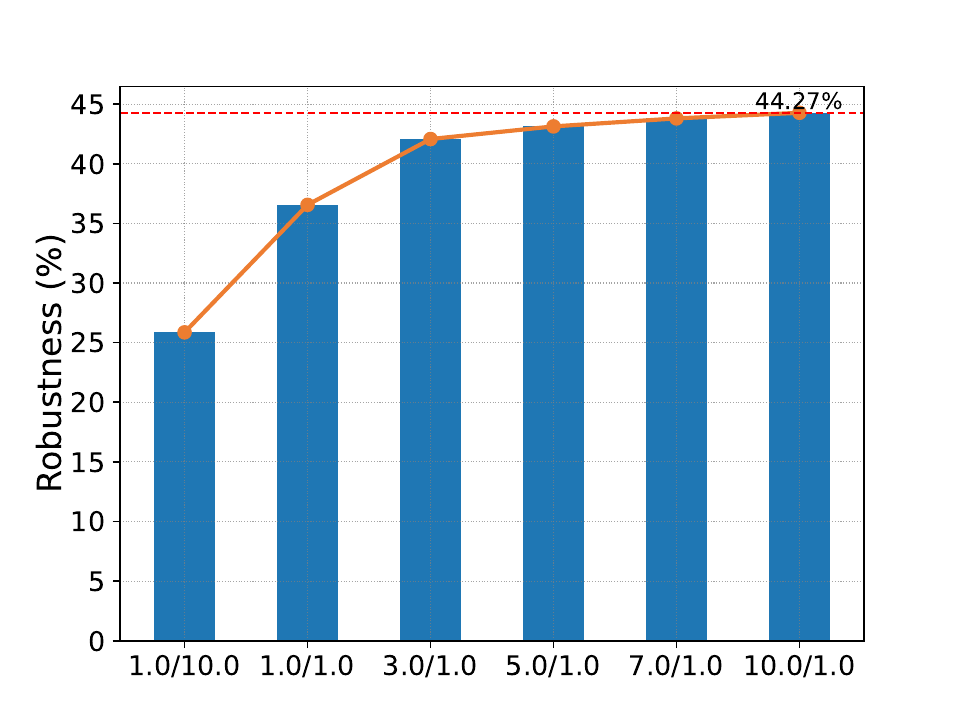}
         (a) MNIST
    \end{minipage}
    \begin{minipage}[b]{0.32\textwidth}
        \centering
         \includegraphics[width=\textwidth]{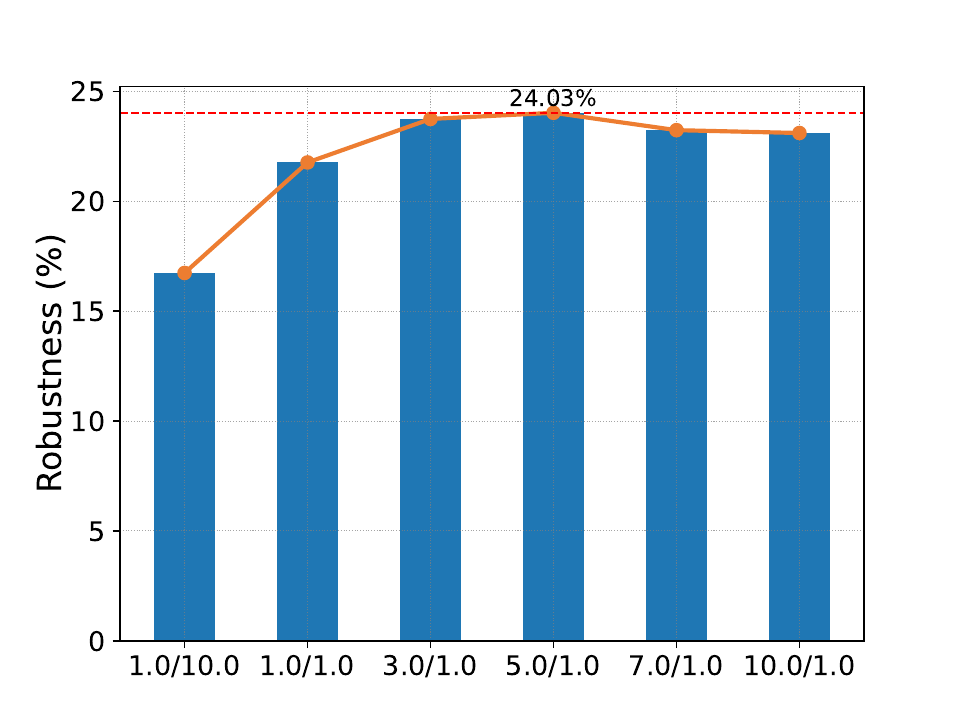}
         (b) CIFAR-10
    \end{minipage}
    \begin{minipage}[b]{0.32\textwidth}
        \centering
         \includegraphics[width=\textwidth]{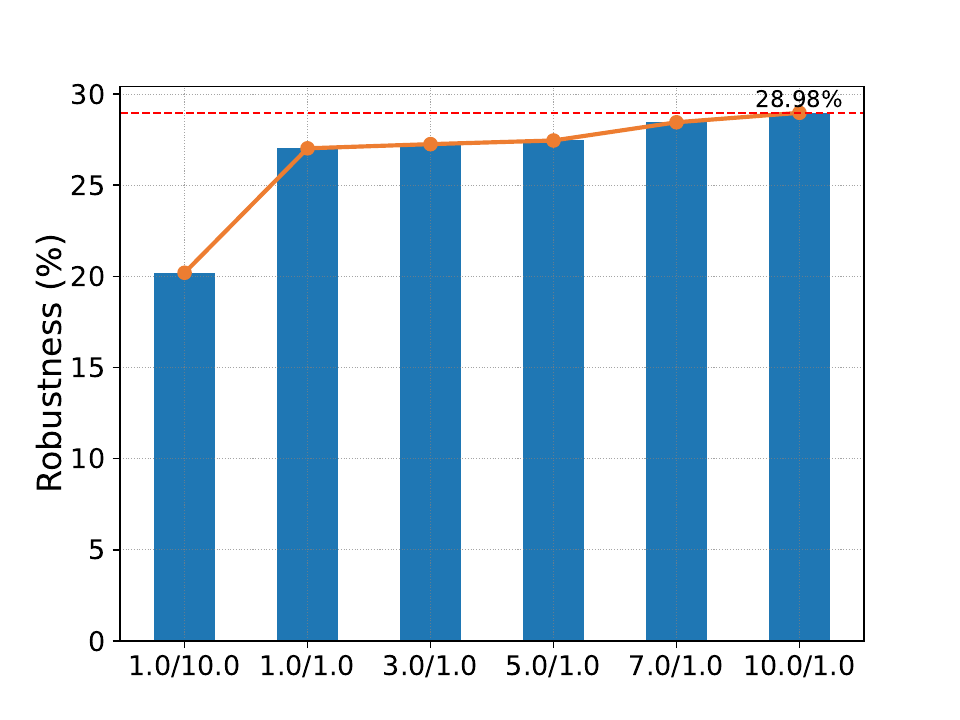}
         (c) CIFAR-100
    \end{minipage}
\caption{Robustness of PM-AFL++ on MNIST, CIFAR-10, and CIFAR-100 under different values of $\rho$. The X-axis represents the ratio of robustness to accuracy, defined as $\rho = \alpha / (1 - \alpha)$. The Y-axis illustrates robustness against a diverse range of attacks. The selected values of $\rho$ for MNIST, CIFAR-10, and CIFAR-100 are 10.0/1.0, 5.0/1.0, and 10.0/1.0, respectively.}
\label{fig:choose_alpha}
\end{figure*}

\textbf{Communication Efficiency.} 
Given that communication has always been a challenge in FL due to the limitations of existing communication channels, we also report the number of communication rounds required for convergence, as well as the number of parameters communicated per round, in Table~\ref{tab:performance_compar}. From the results in the table, it can be observed that the number of parameters communicated per round in PM-AFL++ is significantly lower than that of the other baselines. Moreover, PM-AFL++ requires the fewest communication rounds to complete global model training. For example, in the MNIST results, with approximately 73 times fewer communication parameters per round and 1.6 times fewer communication rounds, PM-AFL++ achieves a clean accuracy of 94.68\% and a robustness accuracy of 44.27\%, both of which are superior to or comparable with other methods. Even in the more challenging task, PM-AFL framework can still reduce the communication parameters per round from 11,690 to 504, which demonstrates a 23 times fewer. Meanwhile, the required communication rounds are also 1.3 times fewer than several baselines. For example, in the MNIST results, with approximately 73 times fewer communication parameters per round and 1.6 times fewer communication rounds, PM-AFL++ achieves a clean accuracy of 94.68\% and a robustness accuracy of 44.27\%, both of which are superior to or comparable with other methods. Even in the more challenging CIFAR-10 task, the PM-AFL framework reduces the communication parameters per round from 11,690 to 320, demonstrating a reduction of 36 times. Additionally, the required communication rounds are 1.3 times fewer than those of several baselines. Overall, these results suggest that our proposal can not only enhance model performance across different data scenarios but also reduce both communication rounds and parameters, demonstrating promising results in achieving communication efficiency in channel-limited edge networks. Therefore, we conjecture that with such a carefully designed framework, comparable or even superior performance can be achieved with fewer resources in real-world, channel-constrained edge network scenarios.

\begin{table}[t]
\centering
\caption{Comparision of distillation temperatures. CA denotes clean accuracy, while RA represents robust accuracy. The optimal trade-off is in \textbf{bold}.}
\label{tab_cifar10_choose_gamma}
\vspace{-5px}
\resizebox{0.99\linewidth}{!}{
  \begin{tabular}{l|cc|cc|cc}
    \toprule
    \multicolumn{1}{l|}{Dataset} & \multicolumn{2}{c|}{MNIST} & \multicolumn{2}{c|}{CIFAR-10} & \multicolumn{2}{c}{CIFAR-100} \\
    \midrule
    $T$ & CA (\%) & RA (\%) & CA (\%) & RA (\%) & CA (\%) & RA (\%) \\
    \midrule
    1.0 & 92.58 & 26.62 & 43.52 & 23.66 & 58.60 & 27.60 \\
    2.0 & 94.62 & 41.87 & \textbf{47.88} & \textbf{24.03}  & \textbf{57.40}& \textbf{28.98} \\
    3.0 & \textbf{94.68} & \textbf{44.27} & 47.08 & 23.25  & 54.60 & 27.63 \\
    4.0 & 94.28 & 43.45 & 46.82 & 23.08  & 54.00 & 26.50 \\
    5.0 & 94.22 & 41.95 & 47.37 & 23.01 & 54.40 & 25.30 \\
    \bottomrule
\end{tabular}}
\end{table}

\textbf{Scalability Comparison.} To provide a more comprehensive evaluation of our proposal across different data heterogeneity scenarios, we conduct a scalability comparison, as shown in Figure~\ref{fig:fgsm_scala_comparision}. This figure illustrates the robustness comparison under the FGSM metric between PM-AFL++ and FedPGD for the MNIST task. The results show that as the parameter $a$ decreases, leading to higher data heterogeneity among clients, the robustness of both PM-AFL++ and FedPGD declines. This indicates that data heterogeneity can affect model performance, with a lower $a$ presenting a greater challenge. However, we observe that PM-AFL++ experiences a slower decline in robustness compared to FedPGD, highlighting its superior adaptability and scalability across various heterogeneity settings. For example, as the data heterogeneity parameter $a$ decreases from 1.0 to 0.1, the robustness of FedPGD drops from 70.64\% to 51.74\%, representing a 26.75\% decline. In contrast, our approach maintains greater stability, with robustness decreasing from 73.32\% to 63.96\%, a more moderate decline of 12.76\%. Similar trends are observed in Figure~\ref{fig:clean_scala_comparision}, which reports the clean accuracy scalability comparison on the more challenging CIFAR-10 task. The results demonstrate that while both methods experience a decline in clean accuracy as data heterogeneity increases, PM-AFL++ maintains higher clean accuracy than FedPGD. For instance, as the data heterogeneity parameter $a$ decreases from 1.0 to 0.1, the accuracy of FedPGD drops from 48.48\% to 28.82\%, representing a 40.55\% decline. In contrast, our approach exhibits greater stability, with accuracy merely decreasing from 50.28\% to 47.88\%, a more moderate decline of 4.77\%. Overall, these results demonstrate that our proposal achieves promising results in both accuracy and robustness when handling varying data distributions.

\begin{table*}[t]
\centering
\caption{Comparison of different methods on benchmark datasets using the same model architecture for local models. The best results are in \textbf{bold} and second with \underline{underline}. PM-AFL and PM-AFL++ outperform the baselines in most cases, with PM-AFL++ achieving higher accuracy (\%) and robustness (\%) while requiring fewer communication rounds.} 
\vspace{-3px}
\label{tab:baselines_ours_same_model}
\setlength{\abovecaptionskip}{0.1cm}
\setlength{\belowcaptionskip}{-0.07cm}
\resizebox{0.95\textwidth}{!}{
\begin{tabular}{llccccccccc}
\toprule
\multirow{2}*{Dataset} & \multirow{2}*{Method} & \multirow{2}*{Clean Acc.} & \multicolumn{7}{c} {Robust Acc.} & \multirow{2}*{\begin{tabular}[c]{@{}c@{}}\# of Comm \\ Rounds \end{tabular}} \\
\cmidrule{4-10}
& & & FGSM & BIM & PGD-40 & PGD-100 & Square & AA & \textbf{Avg} & \\
\midrule
\multirow{10}*{\textbf{MNIST}} & FedAvg~\cite{mcmahan2017communication} & 92.22 & 1.28 & 4.14 & 0.00 & 0.00 & 0.00 & 0.00 & 0.90 &  160   \\
& MixFAT~\cite{zizzo2020fat} & 88.12 & 14.92 & 44.76 & 6.92 & 4.02 & 3.94 & 2.98 & 12.92 &  160   \\
& FedPGD~\cite{madry2018towards} & 87.98 & 16.06 & 47.42 & 7.72 & 4.30 & 4.56 & 3.14 & 13.86 & 160  \\
& FedALP~\cite{kannan2018adversarial} & 86.24 & 24.68 & 52.52 & 13.54 & 8.84 & 8.08 & 6.32 & 18.99 & 180 \\
& FedMART~\cite{wang2019improving} & 85.04 & 22.72 & 51.32 & 13.08 & 9.36 & 7.36 & 6.08 & 18.32 & 160 \\
& FedTRADES~\cite{zhang2019theoretically} & 89.96 & 27.02 & 55.96 & 13.34 & 8.16 & 8.36 & 6.26 & 19.85 & 160 \\
& CalFAT~\cite{chen2022calfat} & 88.64 & 20.52 & 51.20 & 10.14 & 5.78 & 5.94 & 4.16 & 16.29 & 180  \\
& DBFAT~\cite{zhang2023delving} & 91.24 & 37.20 & 62.14 & 18.74 & 10.48 & 10.20 & 8.20 & 24.49 & 180 \\
& PM-AFL (Ours) & \underline{94.53} & \underline{57.89} & \underline{72.15} & \underline{33.41} & \underline{18.72} & \underline{17.84} & \underline{13.66} & \underline{35.61} & \textbf{100}\\
& PM-AFL++ (Ours) & \textbf{94.68} & \textbf{63.96} & \textbf{77.50} & \textbf{43.10} & \textbf{29.92} & \textbf{27.52} & \textbf{23.62} & \textbf{44.27} & \textbf{100} \\
\midrule
\multirow{10}*{\textbf{CIFAR-10}} & FedAvg~\cite{mcmahan2017communication} & 45.26 & 7.62 & 5.40 & 5.42 & 5.35 & 6.50 & 4.28 & 5.76 & 200 \\
& MixFAT~\cite{zizzo2020fat} & 31.88 & 20.32 & 19.52 & 19.62 & 19.60 & 18.28 & 17.32 & 19.11 & 250 \\
& FedPGD~\cite{madry2018towards} & 26.42 & 19.74 & 18.96 & 19.00 & 18.98 & 18.16 & 17.48 & 18.72 & 200 \\
& FedALP~\cite{kannan2018adversarial} & 25.60 & 18.74 & 18.38 & 18.36 & 18.32 & 17.12 & 16.56 & 17.91 &  200 \\
& FedMART~\cite{wang2019improving} & 28.18 & 20.52 & 19.60 & 19.62 & 19.60 & 18.44 & 17.74 & 19.25 & 200 \\
& FedTRADES~\cite{zhang2019theoretically} & 29.76 & 20.64 & 19.66 & 19.70 & 19.66 & 18.02 & 16.92 & 19.10 & 230 \\
& CalFAT~\cite{chen2022calfat} & 26.02 & 18.64 & 17.98 & 17.96 & 17.94 & 17.16 & 16.64 & 17.72 & 200 \\
& DBFAT~\cite{zhang2023delving} & 33.44 & 21.50 & 20.88 & 20.94 & 20.90 & 18.70 & 17.56 & 20.08 &  200 \\
& PM-AFL (Ours) & \underline{45.76} & \underline{24.46} & \underline{23.96} & \underline{22.96} & \underline{22.94} & \underline{21.26} & \underline{18.74} & \underline{22.38} & \textbf{150} \\
& PM-AFL++ (Ours) & \textbf{47.88} & \textbf{26.80} & \textbf{24.62} & \textbf{24.68} & \textbf{24.66} & \textbf{23.20} & \textbf{20.22} & \textbf{24.03} & \textbf{150}\\
\midrule
\multirow{10}*{\textbf{CIFAR-100}} & FedAvg~\cite{mcmahan2017communication} & 54.09 & 1.03 & 0.00 & 0.00 & 0.00 & 0.97 & 0.00 & 0.33 & 200 \\
& MixFAT~\cite{zizzo2020fat} & \underline{54.64} & 24.38 & 20.74 & 20.80 & 20.48 & 20.16 & 17.89 & 20.74 & 200 \\
& FedPGD~\cite{madry2018towards} & 53.40 & 25.01 & 21.74 & 21.80 & 21.48 & 22.16 & 18.40 & 21.76 & 220 \\
& FedALP~\cite{kannan2018adversarial} & 52.26 & 28.40 & 25.09 & 25.83 & 25.70 & 25.60 & \underline{22.16} & 25.46 & 200 \\
& FedMART~\cite{wang2019improving} & 53.60 & 26.63 & 23.80 & 24.78 & 23.80 & 22.12 & 19.43 & 23.42 & 250 \\
& FedTRADES~\cite{zhang2019theoretically} & 53.88 & 28.80 & 22.89 & 23.88 & 23.10 & 23.40 & 20.86 & 23.82 & 200 \\
& CalFAT~\cite{chen2022calfat} & 51.46 & 26.49 & 24.93 & 25.27 & 24.22 & 24.32 & 21.11 & 24.39 & 200 \\
& DBFAT~\cite{zhang2023delving} & 50.80 & 24.33 & 22.62 & 22.78 & 22.60 & 23.44 & 21.23 & 22.83 & 200 \\
& PM-AFL (Ours) & 54.41 & \underline{31.60} & \underline{28.90} & \underline{28.88} & \underline{28.71} & \underline{26.20} & 19.90 &  \underline{27.36} & \textbf{150} \\
& PM-AFL++ (Ours) & \textbf{57.40} & \textbf{33.81} & \textbf{30.22} & \textbf{30.10} & \textbf{30.09} & \textbf{27.60} & \textbf{22.08} & \textbf{28.98} & \textbf{150} \\
\bottomrule
\end{tabular}}
\end{table*}

\subsection{Ablation Study and Analysis}
\textbf{Effects of Key Components.} To thoroughly analyze the effectiveness of each module in our approach, we conduct an ablation study on MNIST, CIFAR-10, and CIFAR-100 to investigate three components: $\mathcal{L}_{VKD}$, $\mathcal{L}_{AKD}$, and $\mathcal{L}_{ALG}$. Quantitative results for these components are presented in Table~\ref{tab:ablation_comparison}. From the results in the table, we have several observations. First, $\mathcal{L}_{VKD}$ significantly improves clean accuracy, with MNIST showing an increase from 91.54\% to 94.34\%. However, it is worth noting that its impact on robust accuracy is relatively limited, yielding only a 2.59\% increase. This aligns with our observation in Table~\ref{tab:motivation}, where $\mathcal{L}_{VKD}$ alone proves insufficient to inherit both accuracy and robustness from the teacher model, underscoring the necessity of additional strategies to effectively defend against adversarial attacks. Second, incorporating $\mathcal{L}_{AKD}$ significantly enhances robustness, with MNIST's robust accuracy rising from 2.59\% to 36.41\%. While its inclusion in CIFAR-10 and CIFAR-100 slightly reduces clean accuracy compared to using only $\mathcal{L}_{VKD}$, it substantially boosts robust accuracy. This reflects the inherent trade-off between accuracy and robustness, where our goal is to enhance robustness while maintaining high clean accuracy. Third, the best trade-off is achieved with the incorporation of $\mathcal{L}_{ALG}$, where clean accuracy improves from 94.40\% to 94.68\% and robust accuracy rises from 36.41\% to 44.27\%. Similar trends are observed in CIFAR-10 and CIFAR-100. These results underscore the essential roles of $\mathcal{L}_{VKD}$, $\mathcal{L}_{AKD}$, and $\mathcal{L}_{ALG}$ in enabling PM-AFL++ to achieve relatively higher clean accuracy and adversarial robustness in the context of AFL.

\textbf{Effects of Weighting Factor.} The weighting factor in~\ref{overall_KD} plays a role in balancing the trade-off between accuracy and robustness. Therefore, we future analyze the impact of the hyperparameter $\alpha$ across different tasks. To quantify this trade-off, we define $\rho = \frac{\alpha}{1 - \alpha}$, which represents the ratio of robustness to accuracy for varying values of $\alpha$. The robustness evaluation results for the three datasets are shown in Figure~\ref{fig:choose_alpha}. To ensure a comprehensive assessment, robustness is measured against a diverse set of adversarial attacks, including FGSM, BIM, PGD-40, PGD-100, Square, and AA attacks. Take the result of CIFAR-10 in Figure~\ref{fig:choose_alpha} (b) as an example, we can observe that as the ratio $\rho$ increases, robustness rises rapidly, reaching a plateau after $\rho = 3.0/1.0$. Beyond this point, when $\rho$ exceeds $3.0/1.0$, robustness fluctuates slightly, with optimal performance observed at $\rho = 5.0/1.0$. Taking the CIFAR-10 results in Figure~\ref{fig:choose_alpha} (b) as an example, we observe that as the ratio $\rho$ increases, robustness improves rapidly, eventually plateauing at $\rho = 5.0/1.0$. Beyond this point, when $\rho$ exceeds $7.0/1.0$, robustness exhibits slight fluctuations, with the optimal performance observed at $\rho = 5.0/1.0$. Similarly, the results in the figure suggest that the optimal performance for both MNIST and CIFAR-100 is achieved at $\rho = 10.0/1.0$.

\textbf{Effects of Temperature.} We conduct ablation studies to investigate the impact of different temperature values $T$ on the distillation process. In general, a higher $T$ results in smoother class probabilities, facilitating the transfer of more information, while a lower $T$ retains sharper distributions with less information distilled~\cite{gou2021knowledge}. Therefore, an appropriate temperature $T$ needs to be carefully selected to balance knowledge transfer and model performance in the distillation process. We report the results for each task across temperature values selected from \{1,2,3,4,5\}, as shown in Table~\ref{tab_cifar10_choose_gamma}. The results indicate that PM-AFL++ achieves optimal trade-off in accuracy and robustness with $T=3$ for MNIST and $T=2$ for both CIFAR-10 and CIFAR-100 tasks.

\subsection{Further Explorations}
To ensure a more comprehensive evaluation of our proposal, we further answer the following key questions:

\textit{How Do Baselines Perform With the Same Model as PM-AFL?} While the PM-AFL framework aims to leverage the teacher model to guide each local model during federated training processes, it remains unclear whether training the baselines from scratch using the same local model as PM-AFL would result in better performance. To explore this, we train all the baselines from scratch using the same model architecture as the PM-AFL framework, with the results presented in Table~\ref{tab:baselines_ours_same_model}. Note that since all baselines, including ours, adopt the same model parameters, they share identical communication costs. Therefore, we omit the column for the number of communication parameters in Table~\ref{tab:baselines_ours_same_model}. The results in the table demonstrate that PM-AFL and PM-AFL++ outperform the baselines in most cases, even with fewer communication rounds. For instance, in the CIFAR-10 task, FedPGD achieves a clean accuracy of 26.42\% and an average robust accuracy of 18.72\%, whereas PM-AFL++ significantly enhances these metrics to 47.88\% and 24.03\%, respectively. Therefore, these results further highlight the superiority of the proposed PM-AFL framework over training from scratch.

\begin{table}[t]
\centering
\caption{Comparison of different model sizes for FedPGD on CIFAR-10 dataset under non-IID data. "WRN-34-10" refers to the WideResNet-34-10 model. All methods are conducted over 200 global iterations.} 
\label{tab:baselines_different_model}
\vspace{-5px}
\setlength{\abovecaptionskip}{0.1cm}
\setlength{\belowcaptionskip}{-0.07cm}
\resizebox{0.49\textwidth}{!}{
\begin{tabular}{lccc}
\toprule
\multirow{2}*{Model} & \multirow{2}*{Clean Acc. (\%)} & \multirow{2}*{AA Acc. (\%)} & \multirow{2}*{\begin{tabular}[c]{@{}c@{}}\# of Comm \\ Params ($\times 10^3$) \end{tabular}} \\
&  &  & \\
\midrule
CNN & 26.42 & 17.48 & 320 \\
ResNet-10 & 33.28 & 19.24 & 4,903 \\
ResNet-12 & 30.18 & 17.74 & 4,977 \\
ResNet-18 & 28.82 & 17.22 & 11,690  \\
ResNet-20 & 27.52 & 16.84 & 17,297 \\
ResNet-34 & 25.02 & 16.10 & 21,282 \\
WRN-34-10 & 27.92 & 16.42 & 48,263\\
\bottomrule
\end{tabular}}
\vspace{-3px}
\end{table}

\begin{table}[t]
\centering
\caption{Comparison of different model sizes for PM-AFL++ on CIFAR-10 dataset under non-IID data. "WRN-34-10" refers to the WideResNet-34-10 model. All methods are conducted over 200 global iterations.} 
\vspace{-5px}
\label{tab:ours_different_model}
\setlength{\abovecaptionskip}{0.1cm}
\setlength{\belowcaptionskip}{-0.07cm}
\resizebox{0.49\textwidth}{!}{
\begin{tabular}{lccc}
\toprule
\multirow{2}*{Model} & \multirow{2}*{Clean Acc. (\%)} & \multirow{2}*{AA Acc. (\%)} & \multirow{2}*{\begin{tabular}[c]{@{}c@{}}\# of Comm \\ Params ($\times 10^3$) \end{tabular}} \\
&  &  & \\
\midrule
CNN & 50.06 & 20.28 & 320 \\
ResNet-10 & 57.98 & 24.66 & 4,903 \\
ResNet-12 & 58.12 & 25.40 & 4,977 \\
ResNet-18 & 57.22 & 25.72 & 11,690  \\
ResNet-20 & 55.28 & 24.90 & 17,297 \\
ResNet-34 & 55.68 & 24.76 & 21,282 \\
WRN-34-10 & 55.94 & 24.78 & 48,263\\
\bottomrule
\end{tabular}}
\vspace{-7px}
\end{table}

\textit{Can Larger Models Benefit Baselines More?} Although the proposed training framework outperforms the baseline, it remains unclear whether the baseline could benefit more from larger models. To address this, we retrain the baseline using different model architectures. Here, we choose FedPGD as the baseline for analysis due to its direct extension from traditional FL to AFL and its widespread adoption~\cite{chen2022calfat,zhang2023delving,qiao2024logit}. Following most studies~\cite{gowal2021improving,huang2023revisiting} that perform robustness analysis on CIFAR-10, we also conduct experiments on this dataset, with the results reported in Table~\ref{tab:baselines_different_model}. From the results, we observe that increasing the model size from CNN to ResNet-10 or ResNet-12 leads to improvements in both clean accuracy and adversarial robustness. However, as the model size continues to grow with architectures like ResNet-18 and WideResNet-34-10 (WRN-34-10), performance declines across both metrics. For instance, scaling from ResNet-18 to WRN-34-10 increases the number of parameters by approximately four times, yet both clean accuracy and adversarial robustness remain nearly unchanged. For instance, the clean accuracy reaches 28.82\% for ResNet-18 but drops to 27.92\% for WRN-34-10. In contrast, despite the CNN model having 150 times fewer parameters than WRN-34-10, it achieves comparable performance. Typically, larger models are expected to yield higher accuracy~\cite{he2016deep}. However, our findings reveal that increasing the model size does not necessarily lead to better performance. This counterintuitive result may be attributed to the increased optimization difficulty as model complexity grows, particularly in the context of the AFL scenario. Nevertheless, these results, to some degree, support our motivation that training a large model from scratch in AFL may not always lead to superior outcomes.

\textit{A Larger Model Can Inherit More Performance From the Teacher?}
In knowledge distillation, the capacity of the student model plays a crucial role in determining how effectively it can absorb knowledge from the teacher~\cite{hinton2015distilling}. This raises an important question: In the context of AFL, does increasing the model size lead to greater performance gains when inheriting knowledge from the teacher? To explore this, we conduct experiments with various model sizes, as shown in Table~\ref{tab:ours_different_model}. The results suggest that our approach can benefit from larger model sizes. For instance, using ResNet-10 for distillation leads to higher clean accuracy and adversarial robustness compared to smaller models like CNN. However, it is worth noting an interesting phenomenon similar to the observation in Question 2: performance gains do not scale linearly with model size. For instance, while ResNet-12 achieves slightly better adversarial accuracy than ResNet-10, the significantly larger WRN-34-10 only offers marginal gains over ResNet-18 in both clean and robust accuracy. This may suggest an intriguing finding: while a larger model can enhance the student’s ability to absorb knowledge from the teacher, selecting an excessively large student model may not always be necessary for effective distillation. A moderately sized model may still achieve strong performance, striking a balance between knowledge transfer and model complexity.

\section{Conclusion}
\label{sec:conclusion}
In this paper, we have proposed the pre-trained model-guided adversarial federated learning (PM-AFL) framework to address the challenges of non-IID data and adversarial attacks in the context of AFL. Our findings reveal that neither vanilla knowledge distillation (VKD) nor adversarial knowledge distillation (AKD) alone is sufficient to effectively inherit the clean and robust accuracy from the teacher model. To overcome this limitation, we further introduce PM-AFL++, a novel training paradigm that seamlessly integrates VKD and AKD into a unified framework, enhanced by an image mixture strategy, to facilitate effective knowledge transfer between the teacher model and local models. Moreover, we incorporate a global alignment term to ensure that local updates remain closely aligned with global updates, thereby mitigating the challenges posed by non-IID data distributions. Extensive experiments on MNIST, CIFAR-10, and CIFAR-100 demonstrate that our proposed method not only achieves comparable or superior performance in addressing both adversarial attacks and non-IID challenges compared to several baselines, but also significantly reduces communication costs by approximately 73x, 36x, and 23x per round, respectively.

\bibliographystyle{ieeetr}
\bibliography{bib_global}

\end{document}